\documentclass[letterpaper, 10 pt, conference]{ieeeconf}  

\IEEEoverridecommandlockouts                              

\usepackage{graphics} 
\usepackage{graphicx}
\usepackage{amsmath} 
\usepackage{xcolor}
\usepackage{multirow}
\usepackage{bm}
\usepackage{makecell}
\usepackage{url}
\usepackage{float}
\usepackage{subcaption}

\usepackage[some]{background}
\SetBgScale{1}
\SetBgContents{\parbox{10cm}{%
  \Huge Draft:  \today\\[18cm]\rotatebox{180}{\Huge Draft:  \today}}}
\SetBgColor{red}
\SetBgAngle{270}
\SetBgOpacity{0.2}

\title{\LARGE \bf
Towards More Efficient EfficientDets and Low-Light Real-Time Marine Debris Detection}

\author{Federico Zocco, Ching-I Huang, Hsueh-Cheng Wang, Mohammad Omar Khyam and Mien Van
\thanks{This work was supported by the Natural Environment Research Council, United Kingdom [grant number NE/V008080/1]. \emph{(Corresponding author: Mien Van.)}}
\thanks{Federico Zocco and Mien Van are with the Centre for Intelligent Autonomous Manufacturing Systems, School of Electronics, Electrical Engineering and Computer Science,
        Queen's University Belfast, Northern Ireland, UK.
        {\tt\small email: \{f.zocco, m.van\}@qub.ac.uk}}%
\thanks{Ching-I Huang and Hsueh-Cheng Wang are with the Department of Electrical and Computer Engineering, National Chiao Tung University (NCTU), also with the Institute of Electrical and Control Engineering, and National Yang Ming Chiao Tung University (NYCU), and also with the Pervasive Artificial Intelligence Research (PAIR) Labs, Hsinchu, Taiwan.
        {\tt\small email: cihuang@nctu.edu.tw; hchengwang@g2.nctu.edu.tw}}%
\thanks{Mohammad Omar Khyam is with the School of Engineering and Technology, Central Queensland University, Australia.
			{\tt\small email: m.khyam@cqu.edu.au}}
}

\begin{document}

\maketitle
\thispagestyle{empty}
\pagestyle{empty}


\begin{abstract}
Marine debris is a problem both for the health of marine environments and for the human health since tiny pieces of plastic called ``microplastics'' resulting from the debris decomposition over the time are entering the food chain at any levels. For marine debris detection and removal, autonomous underwater vehicles (AUVs) are a potential solution. In this letter, we focus on the efficiency of AUV vision for real-time and low-light object detection. First, we improved the efficiency of a class of state-of-the-art object detectors, namely EfficientDets, by 1.5\% AP on D0, 2.6\% AP on D1, 1.2\% AP on D2 and 1.3\% AP on D3 without increasing the GPU latency. Subsequently, we created and made publicly available a dataset for the detection of in-water plastic bags and bottles and trained our improved EfficientDets on this and another dataset for marine debris detection. Finally, we investigated how the detector performance is affected by low-light conditions and compared two low-light underwater image enhancement strategies both in terms of accuracy and latency. Source code and dataset are publicly available\footnote{https://github.com/fedezocco/MoreEffEffDetsAndWPBB-TensorFlow}.   

\vspace{0.5cm}
\end{abstract}

\section{INTRODUCTION}
Nowadays, waste management is an issue of increasing concern due to the accumulation of litter in natural environments such as seas and oceans. Every year, between 1.15 and 2.41 million tonnes of plastic waste enter the ocean from rivers \cite{MarineLitterStatistics}. Plastic waste disposed in natural environments releases tiny fractions of material called microplastics that potentially could harm the health of all species, including humans, because they are ingested or breathed at any levels of the food chain \cite{HarmfulForHumanHealth,USGSmicroplastics,EUmicroplastics}.  

\begin{figure*}[t]
\begin{minipage}{\textwidth}
\includegraphics[width=\textwidth]{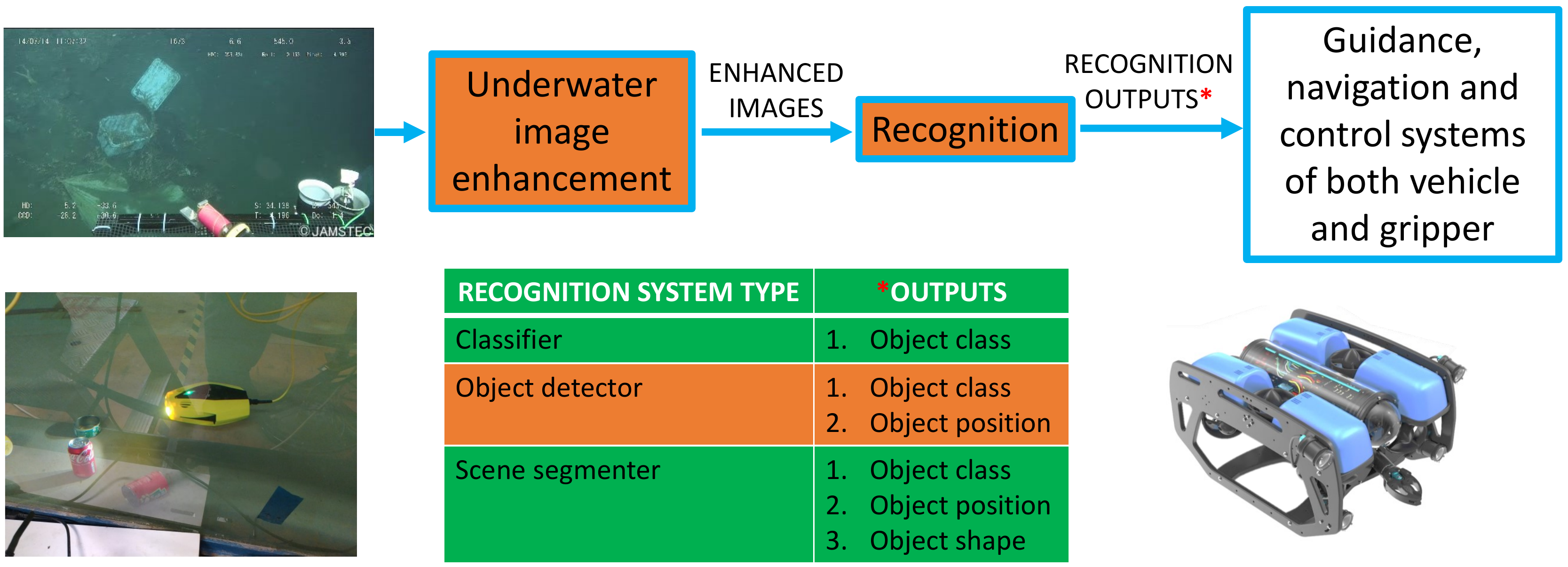}
\centering
\caption[...]{General automation pipeline of an AUV equipped with a gripper for marine debris detection and removal. At the top, from left to right: an underwater image enhancement technique improves the RGB camera images, a recognition algorithm makes the predictions, then the recognition system outputs are used by the guidance, navigation and control systems of both vehicle and gripper. Different recognition systems provide different outputs as detailed in the table. This paper focuses on the orange blocks by seeking the improvement of the efficiency of a set of state-of-the-art object detectors called EfficientDets \cite{EfficientDets} and investigating the efficiency of the whole pipeline with and without low-light underwater image enhancement. Bottom left: we collected images of waste items inside a testing pool using the Chasing Dory drone. Bottom right: the BlueROV2, a commercial robot equipped with a gripper\textsuperscript{a}.}\tiny\textsuperscript{a}BlueROV2 image sourced from: \url{https://free3d.com/3d-model/underwater-robot-bluerov2-rigged-4532.html}
\label{fig:AutomationOfAUV}
\end{minipage}
\end{figure*}

For marine litter removal, an emerging approach consists of deploying fleets of underwater vehicles either fully autonomous or partially operated by cleaning divers. In order to perform debris detection and removal, the automation pipeline of an autonomous underwater vehicle (AUV) could be structured as in Fig. \ref{fig:AutomationOfAUV}: an underwater image enhancement (UIE) block, a recognition block to recognize the class of the objects, their positions, their shapes and constituent materials, and finally the block of guidance, navigation and control to guide the vehicle towards the objects and to grab them leveraging the recognition system outputs \cite{AUVandGrasping}. In this paper, we focus on the UIE and the recognition blocks.

\begin{figure}[H]
\includegraphics[width=0.48\textwidth]{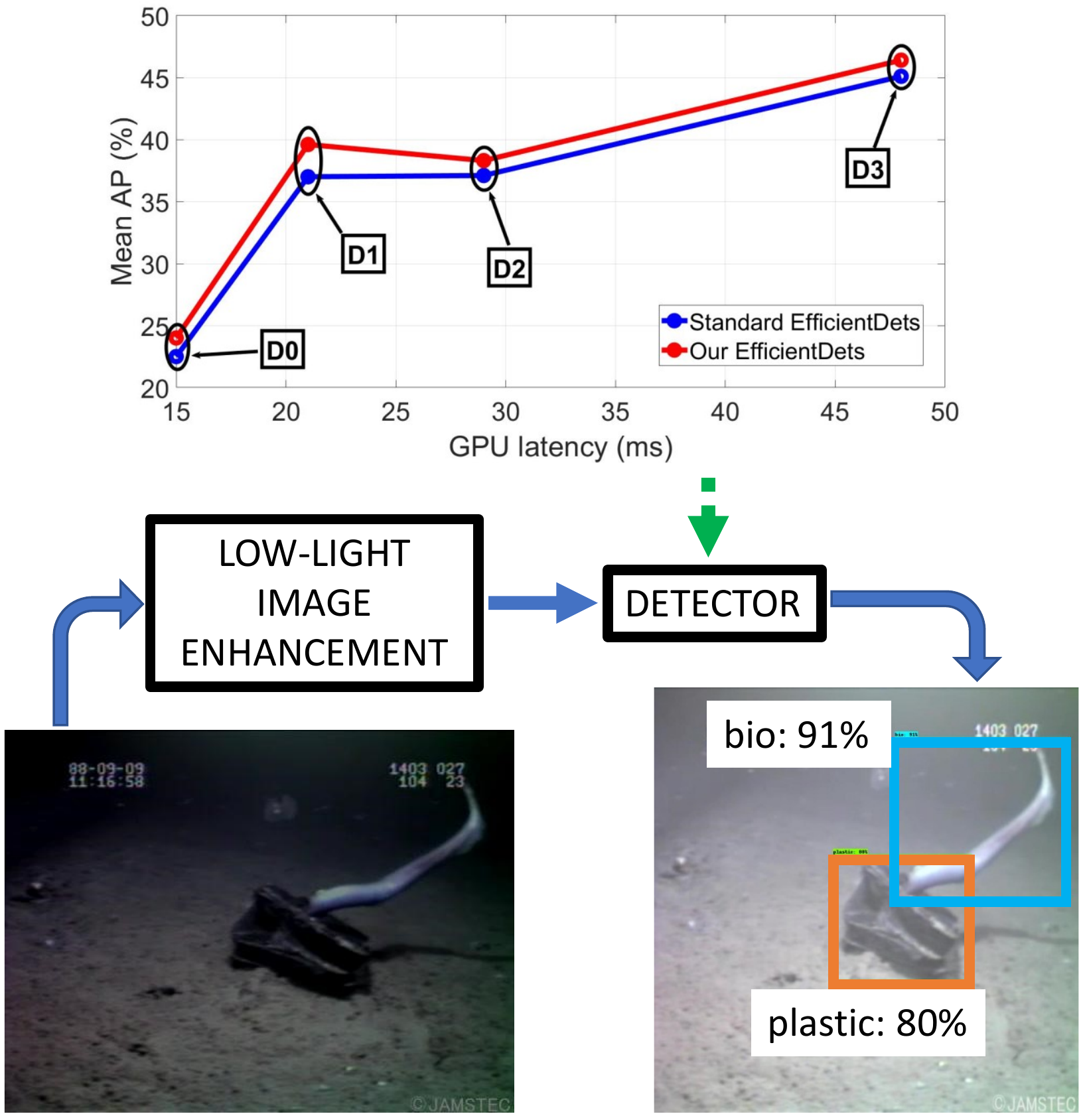}
\centering
\caption{Summary of the paper contributions: we found an architecture for the parent detector D0 that led to more efficient EfficientDets detectors \cite{EfficientDets} on PASCAL VOC 2012. Then, we trained our EfficientDets for marine debris detection and investigated the performance of the low-light image enhancement method proposed in \cite{L2UWE} when combined with our detectors.}
\label{fig:mAPvsLatency}
\end{figure} 

Most of the state-of-the-art computer vision algorithms are based on convolutional neural networks (CNN) and an end-to-end learning paradigm inspired by the biological learning process \cite{LeCunCNN}. The approach of learning the image features is based on the deep learning of representations (typically shortened with ``deep learning'') and is increasingly replacing the classical techniques based on hand-crafted features \cite{SzeliskiBook} such as bag-of-words \cite{ObsoleteCV} and part-based \cite{ObsoleteCV2} models. As the most accurate object detectors have computational and memory costs prohibitive for real-time applications such as marine debris detection and removal, finding the best accuracy-latency compromise is an active area of research \cite{EfficientDets}.     

\textbf{Our contributions.} 
\begin{enumerate}
\item{As the object detectors recently proposed by Tan et al. \cite{EfficientDets} showed state-of-the-art accuracy-latency performance, we investigated two architectural alternatives of their baseline detector D0. On PASCAL VOC 2012, we found that one of our architectures, namely D0(1-5), is 1.5\% more accurate than the one proposed in \cite{EfficientDets} while having comparable latency. Then, by scaling-up D0(1-5), we achieve a more efficient class of EfficientDets as shown in Fig. \ref{fig:mAPvsLatency}.}
\item{We trained D0(1-5) and its derivations D1-D3 on 1200 samples of the Trash-ICRA19 dataset \cite{TrashDataset}. The results are in Table \ref{tab:ResultsWithTrashDataset}.}
\item{Since the Trash-ICRA19 dataset \cite{TrashDataset} does not specify the type of plastic object, we created the in-water plastic bags and bottles (WPBB) dataset and made it publicly available; then we trained D0(1-5) and its derivations on the WPBB dataset. The results are in Table \ref{tab:ResultsWithWPBBDataset} and Fig. \ref{fig:DebrisDetectionsPBBW}.}
\item{Assuming low-light conditions, we address two questions: ``Is it more efficient scaling-up the object detector size (e.g. from D0(1-5) to D3(1-5)) or adding an underwater image enhancement method before the smallest detector D0(1-5)?'' and ``How does the method proposed in \cite{L2UWE} perform as pre-processing step for object detection?''. These investigations are covered in Subsection \ref{sub:UIE}.}
\end{enumerate} 

The rest of the paper is organized as follows: Section \ref{sec:RelWorks} covers the related work, Section \ref{sec:OurChangesToEfficientDets} details our modifications to the class of detectors proposed by Tan et al. \cite{EfficientDets}, Section \ref{sec:Experiments} presents and discusses the experimental results and finally Section \ref{sec:Conclusions} concludes.     

\section{RELATED WORK}
\label{sec:RelWorks}
\subsection{Underwater Image Enhancement}
As the images recorded in-water often suffer from low-light conditions, light refraction, light absorption, scattering or water turbidity, several works focused on improving the quality of underwater images \cite{imageEnhancement1,imageEnhancement2,imageEnhancement3,FUIE,L2UWE}, with \cite{imageEnhancement1,imageEnhancement2,imageEnhancement3,FUIE} based on generative adversarial networks \cite{GANpaper}. The authors of \cite{imageEnhancement1} developed both a GAN and a CNN. The former is to generate underwater synthetic images from images taken in air with the aim of generating a large amount of in-water data to train deep image enhancement models. The latter is an underwater image enhancement model trained using the GAN synthetic outputs. One of the very first works proposing a GAN to improve in-water images is \cite{imageEnhancement2}, where a lack of distorted in-water images was addressed by using CycleGAN \cite{CycleGAN} to generate distorted images from undistorted ones; then, the paired dataset is used to train a generator that performs the opposite mapping of CycleGAN, i.e. it outputs the undistorted version of a distorted input image. Guo et al. \cite{imageEnhancement3} developed a densely connected generator by introducing a multiscale dense block inside a fully convolutional architecture. This choice was inspired by DenseNet \cite{DenseNet}, where a large number of skip-connections strengthen feature reuse and feature propagation leading to an improvement in CNN efficiency. A different GAN was designed in \cite{FUIE} for underwater image enhancement: the generator architecture has five encoder-decoder pairs with mirrored skip-connections, whereas the discriminator is a Markovian PatchGAN \cite{PatchGAN}; the loss function was specifically defined to take into account the aspects of global similarity, image content and local textures. A non-adversarial approach is \cite{L2UWE}, where the authors generated two lighting distribution models for local and non-local regions, respectively. This resulted in two different dehazing maps which subsequently are combined into a single output that preserves the image details while removing the darkness from the original image.

\subsection{Efficiency of Object Detectors}
The last years have seen several works addressing both the speed and the accuracy of object detection to permit the use of detectors in edge devices and real-time applications. In 2016, two of the first works in this direction were \cite{Ref24OfTan}, which removed the proposal generation and re-sampling stages, and \cite{Ref30OfTan}, which added batch normalization to all convolutional layers and removed dropout from its model. In 2017, Lin et al. \cite{Ref21OfTan} boosted the detection efficiency by modifying the cross entropy loss typically used for the classification subnet. In 2018, Law et al. \cite{Ref18OfTan} replaced the prediction of anchor boxes with pair of keypoints, while Huang et al. \cite{Ref26OfTan} opted for removing some layers and the batch normalization operation to speed-up a larger detector; in contrast, \cite{Ref31OfTan} designed a larger backbone network to improve the accuracy of a previous smaller model. In 2019, Zhou et al. \cite{Ref41OfTan} proposed to predict center points of each object as keypoints and regressing width and high of the corresponding bounding box, while Tian et al. \cite{Ref37OfTan} simplified the detection framework by formulating the problem as a per-pixel prediction analogously to instance segmentation. More recently, Tan et al. \cite{EfficientDets} introduced a bidirectional feature pyramid network for fast multi-scale feature fusion, while Cai et al. \cite{AnotherDetectorSeekingEfficiency} simplified a larger detector through an inter-block weight pruning of convolutional layers, where a block contains the weights of $n$ consecutive channels of $m$ filters.

\section{VARIANTS OF EfficientDets}
\label{sec:OurChangesToEfficientDets}
\subsection{Original Architecture}
Tan et al. \cite{EfficientDets} developed a set of eight detectors named EfficientDets obtained by scaling-up the parent architecture EfficientDet D0. The detectors share the same general structure: they have an EfficientNet \cite{ScalingUpCNNs} as the backbone network, which is then followed by one or more bi-directional feature pyramid network (BiFPN) layers and end with two separate identical sub-nets, one for class prediction and one for box prediction. 

\noindent
\textbf{EfficientNets.} It is becoming more common designing a CNN of small size and derive a set of larger child networks from it by scaling-up its dimensions. One of the first systematic studies of model scaling for CNN was \cite{ScalingUpCNNs}, where a set of six classifiers, namely EfficientNets, were developed by scaling-up a small baseline CNN, namely EfficientNet-B0. The architecture of B0 resulted from a multi-objective neural architecture search inspired by \cite{NeuralArchitectureSearch2}. While previous works scaled-up single dimensions, e.g. the depth \cite{ScalingUpByDepth} or the width \cite{ScalingUpByWidth}, Tan et al. \cite{ScalingUpCNNs} jointly increased depth, width and resolution by a constant ratio and empirically showed the benefit in terms of efficiency given by jointly scaling all the network dimensions.          

\noindent
\textbf{BiFPN.} The BiFPN layer was proposed in \cite{EfficientDets} to merge features from different scales generated through a pyramidal feature hierarchy. It can be seen as the successor of three previous approaches for cross-scale feature fusion: FPN \cite{ForBiFPN1} proposed a top-down pathway on top of a bottom-up pyramid feature hierarchy; then PANet \cite{ForBiFPN2} added a further bottom-up layer that merges features from both FPN pathways; finally \cite{ForBiFPN3} used the neural architecture search algorithm \cite{NeuralArchitectureSearch} to define an architectural topology of a block for cross-scale feature representation fusion whose copies can be stacked multiple times. BiFPN takes the top-down pathway from FPN, the extra bottom-up layer from PANet and the use of a repeatable block from \cite{ForBiFPN3}. While previous methods treated equally the features from different scales, another peculiarity of BiFPN is that cross-scale features are weighed depending on their resolution \cite{EfficientDets}.

\noindent
\textbf{Detector Scaling.} Scaling-up object detectors is typically based on using a bigger backbone network or stacking more FPN layers \cite{ForBiFPN3}. An object detector has generally more dimensions than a classifier since it extends the prediction capability of the latter, hence the grid search done in \cite{ScalingUpCNNs} to define the scaling factors was replaced in \cite{EfficientDets} by an heuristic-based approach: a coefficient $\phi$ that determines at the same time the size of the backbone network, the resolution, the number of channels and the number of layers of the network building blocks. The rationale behind a uniform scaling across dimensions is that an higher resolution image requires deeper and wider architectures to capture the information contained in more pixels. This intuition was empirically supported in \cite{ScalingUpCNNs} as the increase of width without changing depth and resolution led to faster accuracy saturation compared to an all-dimension scaling.

\vspace{0.5cm}    
The parent EfficientDet is D0, it is the smallest detector and made of 3 BiFPN layers, 3 convolutions in class/box sub-nets and B0 as the backbone. The other EfficientDets, namely D1-D7, result from scaling-up D0 according to the scaling factor $\phi = \{0, 1, \dots, 7\}$. The architecture size is augmented by jointly increasing all the dimensions: the input image resolution ($R^{in}$), the backbone network, the number of channels of the BiFPN layer and class/box nets ($W^{Bi,cb}$), the number of BiFPN layers ($D^{Bi}$) and the depth of the class/box prediction sub-nets ($D^{cb}$). Specifically, the re-scaling equations in Tan et al. \cite{EfficientDets} are              
\begin{equation}
W^{Bi,cb} = 64 \cdot (1.35^\phi),
\label{eq:TanWidth}
\end{equation}
\begin{equation}
D^{Bi} = 3 + \phi,
\label{eq:TanDepth1}
\end{equation}
\begin{equation}
D^{cb} = 3 + \left \lfloor{\phi / 3}\right \rfloor
\label{eq:TanDepth2}
\end{equation}
and
\begin{equation}
R^{in} = 512 + \phi \cdot 128.
\label{eq:TanResolution}
\end{equation}
Since D0 corresponds to $\phi = 0$, from (\ref{eq:TanDepth1}) and (\ref{eq:TanDepth2}) follows that the parent network has 3 BiFPN layers and 3 convolutions in both class and box prediction nets, i.e. $D^{Bi} = D^{cb} = 3$.  

\subsection{Our Architectural Modifications}
\begin{figure*}[t]
\includegraphics[width=\textwidth]{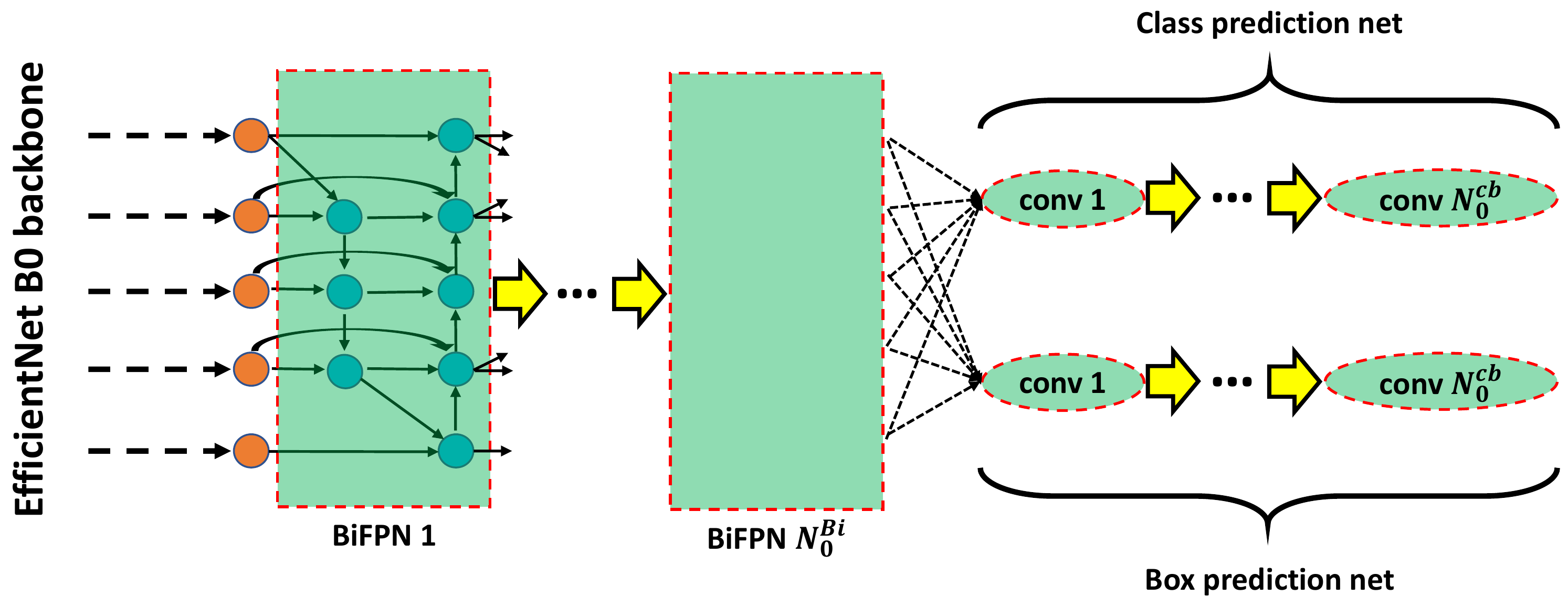}
\centering
\caption{Our architectural modifications of EfficientDet D0. The number of BiFPN layers (i.e. $N^{Bi}_0$) and the number of convolutional layers in class/box nets (i.e. $N^{cb}_0$) proposed in \cite{EfficientDets} for D0 are 3 and 3, respectively. We keep $N^{Bi}_0 + N^{cb}_0 = 6$ and change the layer distribution: ($N^{Bi}_0$; $N^{cb}_0$) = (1; 5) for the architecture D0(1-5) and ($N^{Bi}_0$; $N^{cb}_0$) = (5; 1) for the architecture D0(5-1).}
\label{fig:OurArchitectures}
\end{figure*}

We observe that the design of D0 is particularly important as its efficiency affects the one of all the child detectors D1-D7 \cite{ScalingUpCNNs}. Hence, we investigate the performance of two variants of D0 by modifying (\ref{eq:TanDepth1}) and (\ref{eq:TanDepth2}) as 
\begin{equation}
D^{Bi} = N^{Bi}_0 + \phi
\end{equation}
and
\begin{equation}
D^{cb} = N^{cb}_0 + \left \lfloor{\phi / 3}\right \rfloor,
\end{equation}
respectively. Here, $N^{Bi}_0$ is the number of BiFPN layers in D0 and $N^{cb}_0$ is the number of convolutional layers for box/class prediction nets in D0. While Tan et al. \cite{EfficientDets} set $(N^{Bi}_0$; $N^{cb}_0) = (3; 3)$, to keep approximately constant the complexity of any D0 variants we fix the sum $N^{Bi}_0 + N^{cb}_0 = 6$, then we investigate an architecture with ($N^{Bi}_0$; $N^{cb}_0$) = (1; 5) and one with ($N^{Bi}_0$; $N^{cb}_0$) = (5; 1) to see \emph{whether is more efficient having more BiFPN layers or more convolutions in the class/box subnets}. Figure \ref{fig:OurArchitectures} clarifies our modifications of D0, while Table \ref{tab:netsScaledFrom1-5} and Table \ref{tab:netsScaledFrom5-1} detail the scaling configurations of the detectors D0-D3 obtained by scaling-up D0(1-5) and D0(5-1), respectively.        
\begin{table}
\centering
\caption{Scaling configurations for EfficientDets D0-D3 considering D0(1-5) as the parent architecture. The architecture D$n$(1-5) is scaled considering $\phi = n$ in (\ref{eq:TanWidth}), (\ref{eq:TanDepth1}), (\ref{eq:TanDepth2}) and (\ref{eq:TanResolution}). The values in bold differ from \cite{EfficientDets} because influenced by our modifications.}
 \begin{tabular}{cccccc} 
 \hline
Architecture & \makecell{Input size \\ ${R^{in}}$} & \makecell{Backbone \\ network} & $W^{Bi,cb}$ & $D^{Bi}$ & $D^{cb}$\\  
\hline
D0(1-5) & 512 & B0 & 64 & \textbf{1} & \textbf{5}\\ 
D1(1-5) & 640 & B1 & 88 & \textbf{2} & \textbf{5}\\ 
D2(1-5) & 768 & B2 & 112 & \textbf{3} & \textbf{5}\\ 
D3(1-5) & 896 & B3 & 160 & \textbf{4} & \textbf{6}\\ 
 \hline
 \end{tabular}
\label{tab:netsScaledFrom1-5}
\end{table}
\begin{table}
\centering
\caption{Scaling configurations for EfficientDets D0-D3 considering D0(5-1) as the parent architecture. The architecture D$n$(5-1) is scaled considering $\phi = n$ in (\ref{eq:TanWidth}), (\ref{eq:TanDepth1}), (\ref{eq:TanDepth2}) and (\ref{eq:TanResolution}). The values in bold differ from \cite{EfficientDets} because influenced by our modifications.}
 \begin{tabular}{cccccc} 
 \hline
Architecture & \makecell{Input size \\ ${R^{in}}$} & \makecell{Backbone \\ network} & $W^{Bi,cb}$ & $D^{Bi}$ & $D^{cb}$\\  
\hline
D0(5-1) & 512 & B0 & 64 & \textbf{5} & \textbf{1}\\ 
D1(5-1) & 640 & B1 & 88 & \textbf{6} & \textbf{1}\\ 
D2(5-1) & 768 & B2 & 112 & \textbf{7} & \textbf{1}\\ 
D3(5-1) & 896 & B3 & 160 & \textbf{8} & \textbf{2}\\   
 \hline
 \end{tabular}
 \label{tab:netsScaledFrom5-1}
\end{table}

\section{EXPERIMENTS}
\label{sec:Experiments}
Any training covered in this letter were performed on Google Colaboratory, which randomly assigned us either an Nvidia Tesla P100 or an Nvidia Tesla T4 GPU. Shorter and longer trainings were performed with approximately 12 GB and 25 GB of RAM, respectively. We used SGD as optimizer and the EfficientNet backbones were initialized with ImageNet checkpoints; then the whole networks were trained on each target dataset: PASCAL VOC 2012 in Subsection \ref{sub:EffDetsBenchmarking} for benchmarking, Trash-ICRA19 and WPBB in Subsection \ref{sub:TrashAndWPBB} for marine debris detection. Most of the source code for this paper is a modification of the original TensorFlow implementation of EfficientDets\footnote{https://github.com/google/automl/tree/master/efficientdet}. We also used the MATLAB implementation of $\text{L}^2$UWE\footnote{https://github.com/tunai/l2uwe} for Subsection \ref{sub:UIE}.

\subsection{Comparison of Original and Our EfficientDets}\label{sub:EffDetsBenchmarking}
Given that
\begin{enumerate}
\item{the training of EfficientDets on PASCAL VOC 2012 is time consuming,}
\item{the neural network training is not deterministic due to the random weight initialization,} 
\end{enumerate}
initially we trained for 5 times the three variants of D0 (i.e. D0(3-3) from \cite{EfficientDets}, D0(1-5) and D0(5-1)) considering just 15 epochs. The results are shown in Table \ref{tab:Selection}. As D0(5-1) was the least accurate with a latency similar to the other candidates, we discarded D0(5-1) and re-trained the remaining detectors with a larger number of epochs, 25 precisely. The results are shown in Table \ref{tab:D0s}.     
\begin{table}
\centering
\caption{Accuracy and latency of the candidate variants of D0 trained for 15 epochs on PASCAL VOC 2012. The values reported are the best in 15 epochs (not necessarily at the last epoch).}
 \begin{tabular}{c@{\hskip 0.1in}c@{\hskip 0.1in}c@{\hskip 0.1in}c@{\hskip 0.1in}c|cc} 
 \hline
 Training & Architecture & \multicolumn{3}{c}{Accuracy (\%)} & \multicolumn{2}{c}{Latency (ms)} \\
 & & AP & AP50 & AP75 & GPU & CPU \\ 
 \hline
 \multirow{3}{*}{1} & D0(3-3) & 11.2 & 21.9 & 10.4 & \multirow{12}{*}{D0(3-3)} & \multirow{12}{*}{D0(3-3)} \\
                             & D0(1-5) & 11.5 & 21.9 & 11.0 & \multirow{12}{*}{15} & \multirow{12}{*}{186} \\
                             & D0(5-1) & 9.2 & 17.9 & 8.6 & \multirow{14}{*}{D0(1-5)} & \multirow{14}{*}{D0(1-5)} \\
                             \cline{1-5}
\multirow{3}{*}{2} & D0(3-3) & 10.0 & 19.6 & 9.2 & \multirow{14}{*}{15} & \multirow{14}{*}{179}\\
                             & D0(1-5) & 11.1 & 21.8 & 10.1 & \multirow{16}{*}{D0(5-1)} & \multirow{16}{*}{D0(5-1)}\\
                             & D0(5-1) & 7.7 & 15.3 & 6.8 & \multirow{16}{*}{16}& \multirow{16}{*}{209}\\
                             \cline{1-5}
\multirow{3}{*}{3} & D0(3-3) & 10.8 & 21.1 & 9.9 &&\\
                             & D0(1-5) & 9.6 & 18.8 & 8.8 &&\\
                             & D0(5-1) & 9.4 & 18.5 & 8.7 &&\\
                             \cline{1-5}
\multirow{3}{*}{4} & D0(3-3) & 11.7 & 22.7 & 10.8 &&\\
                             & D0(1-5) & 11.9 & 23.1 & 11.0 &&\\
                             & D0(5-1) & 8.4 & 16.9 & 7.6 &&\\
                             \cline{1-5}
\multirow{3}{*}{5} & D0(3-3) & 11.7 & 23.4 & 10.6 &&\\
                             & D0(1-5) & 12.2 & 24.0 & 11.3 &&\\
                             & D0(5-1) & 9.8 & 19.6 & 8.7 &&\\
\cline{1-5} 
\multirow{3}{*}{Mean} & D0(3-3) & 11.1 & 21.7 & 10.2 &&\\
                                 & \textbf{D0(1-5)} & $\bm{11.3}$ & $\bm{21.9}$ & $\bm{10.4}$ &&\\
                                 & D0(5-1) & 8.9 & 17.6 & 8.1 &&\\ 
 \hline
 \end{tabular}
\label{tab:Selection}
\end{table}
\begin{table}
\centering
\caption{Accuracy and latency of the candidate variants of D0 trained for 25 epochs on PASCAL VOC 2012. The values are the best in 25 epochs (not necessarily at the last epoch).}
 \begin{tabular}{c@{\hskip 0.1in}c@{\hskip 0.1in}c@{\hskip 0.1in}c@{\hskip 0.1in}c|cc} 
 \hline
 Training & Architecture & \multicolumn{3}{c}{Accuracy (\%)} & \multicolumn{2}{c}{Latency (ms)} \\
 & & AP & AP50 & AP75 & GPU & CPU \\
 \hline
 \multirow{2}{*}{1} & D0(3-3) & 19.9 & 35.8 & 20.5 & \multirow{8}{*}{D0(3-3)} & \multirow{8}{*}{D0(3-3)}\\
                             & D0(1-5) & 23.5 & 41.4 & 24.2 & \multirow{8}{*}{15} & \multirow{8}{*}{186}\\
                              \cline{1-5}
\multirow{2}{*}{2} & D0(3-3) & 23.9 & 44.2 & 23.5 & \multirow{10}{*}{D0(1-5)} & \multirow{10}{*}{D0(1-5)}\\
                             & D0(1-5) & 23.7 & 42.6 & 24.3 & \multirow{10}{*}{15} & \multirow{10}{*}{179}\\
                              \cline{1-5}
\multirow{2}{*}{3} & D0(3-3) & 24.4 & 43.3 & 25.6 &&\\
                             & D0(1-5) & 23.9 & 42.8 & 24.5 &&\\
                              \cline{1-5}
\multirow{2}{*}{4} & D0(3-3) & 22.0 & 40.0 & 22.4 &&\\
                             & D0(1-5) & 25.2 & 44.8 & 25.7 &&\\
                              \cline{1-5}
\multirow{2}{*}{5} & D0(3-3) & 22.5 & 40.7 & 23.0 &&\\
                             & D0(1-5) & 23.6 & 41.3 & 24.9 &&\\
                             \cline{1-5}
\multirow{2}{*}{Mean} & D0(3-3) & 22.5 & 40.8 & 23.0 &&\\
                                  & \textbf{D0(1-5)} & \textbf{24.0} & \textbf{42.6} & \textbf{24.7} &&\\ 
 \hline
 \end{tabular}
\label{tab:D0s}
\end{table}
Both Table \ref{tab:Selection} and Table \ref{tab:D0s} show that \emph{the detector with deeper class/box subnets and lowest number of BiFPN layers}, i.e. D0(1-5), \emph{is the most accurate}. Therefore, we have deduced that it was more efficient to have less BiFPN layers and that the detector D0(3-3) proposed in \cite{EfficientDets} has an intermediate performance between D0(1-5) and D0(5-1).

Subsequently, we scaled-up the parent detectors D0(1-5) and D0(3-3) considering $\phi = 3$ to get D3(1-5) and D3(3-3), respectively, and performed 3 trainings with 29 epochs. As visible in Table \ref{tab:D3s}, D3(1-5) is 1.3\% more accurate than D0(3-3) on average while having a comparable latency. These results further supported that \emph{more convolutions in the class/box subnets are more efficient than more BiFPN layers}.     
\begin{table}
\centering
\caption{Accuracy and latency of the candidate variants of D3 trained for 29 epochs on PASCAL VOC 2012. The values are the best in 29 epochs (not necessarily at the last epoch).}
 \begin{tabular}{c@{\hskip 0.1in}c@{\hskip 0.1in}c@{\hskip 0.1in}c@{\hskip 0.1in}c|cc} 
 \hline
 Training & Architecture & \multicolumn{3}{c}{Accuracy (\%)} & \multicolumn{2}{c}{Latency (ms)} \\
 & & AP & AP50 & AP75 & GPU & CPU \\
 \hline
 \multirow{2}{*}{1} & D3(3-3) & 46.0 & 74.0 & 50.4 & \multirow{4}{*}{D3(3-3)} & \multirow{4}{*}{D3(3-3)}\\
                             & D3(1-5) & 47.2 & 74.7 & 52.3 & \multirow{4}{*}{48} & \multirow{4}{*}{1337}\\
                              \cline{1-5}
\multirow{2}{*}{2} & D3(3-3) & 45.0 & 72.5 & 49.5 & \multirow{6}{*}{D3(1-5)} & \multirow{6}{*}{D3(1-5)}\\
                             & D3(1-5) & 46.5 & 74.1 & 51.5 & \multirow{6}{*}{48} & \multirow{6}{*}{1423}\\
                              \cline{1-5}
\multirow{2}{*}{3} & D3(3-3) & 44.3 & 71.7 & 49.0 &&\\
                             & D3(1-5) & 45.6 & 72.9 & 51.3 &&\\
                              \cline{1-5}
\multirow{2}{*}{Mean} & D3(3-3) & 45.1 & 72.7 & 49.6 &&\\
                                  & D3(1-5) & \textbf{46.4} & \textbf{73.9} & \textbf{51.7} &&\\ 
 \hline
 \end{tabular}
\label{tab:D3s}
\end{table} 

Then, we considered $\phi = 1$ and $\phi = 2$ to compare the detectors D1 and D2 and the results are reported in Table \ref{tab:D1andD2}. A summary of the comparison between the standard EfficientDets achieved by scaling-up D0(3-3) and our variants achieved by scaling-up D0(1-5) is depicted in Fig. \ref{fig:mAPvsLatency} where it is visible the efficiency improvement with less BiFPN layers.
\begin{table}
\centering
\caption{Accuracy and latency of the candidate variants of D1 and D2 on PASCAL VOC 2012. The values are the best in the total number of epochs (not necessarily at the last epoch).}
 \begin{tabular}{c@{\hskip 0.1in}c@{\hskip 0.1in}c@{\hskip 0.1in}c@{\hskip 0.1in}c|cc} 
 \hline
 Architecture & \# epochs & \multicolumn{3}{c}{Accuracy (\%)} & \multicolumn{2}{c}{Latency (ms)} \\
 & & AP & AP50 & AP75 & GPU & CPU \\
 \hline
 D1(3-3) & 35 & 37.0 & 62.9 & 39.7 & 21 & 394\\
 D1(1-5) & 35 & \textbf{39.6} & \textbf{65.6} & \textbf{42.9} & 21 & 395\\
\hline
 D2(3-3) & 29 & 37.1 & 62.9 & 40.1 & 29 & 656\\
 D2(1-5) & 29 & \textbf{38.3} & \textbf{64.6} & \textbf{41.9} & 29 & 652\\
 \hline
 \end{tabular}
\label{tab:D1andD2}
\end{table}

\subsection{On Marine Debris Datasets}\label{sub:TrashAndWPBB}
Subsequently, we trained our improved EfficientDets on the first 1200 samples of the Trash-ICRA19 \cite{TrashDataset}, which is a publicly available annotated dataset for object detection targeting three classes: ``bio'' for biological items, ``plastic'' for plastic items and ``rov'' for remotely operated vehicles. The results are in Table \ref{tab:ResultsWithTrashDataset}, where 35 epochs have been chosen for demonstration purposes; when training a model for deployment on an AUV, the number of epochs should be increased in order to reach the learning saturation. With 35 epochs, we reached 47.4\% AP with D0 and 49.6\% AP with D3. In terms of latency, there is a significant difference between the use of a CPU and a GPU. On a CPU, all detectors take more than 0.1 seconds and D3 requires more than 1 second to process a single image. A GPU on-board would certainly speed-up the detection, however the speed-up will likely be smaller than the one reported in \ref{tab:ResultsWithTrashDataset} because the GPU on the AUV will likely be less powerful than Google Colab GPUs (for this work, we were randomly assigned either a Tesla P100 or a Tesla T4). Note that, along with speeding-up the detectors, a GPU reduces also the increase of latency with the scaling-up of the detector size: while on a CPU D1, D2 and D3 are 2.2, 3.6 and 7.9 times slower than D0, respectively, on a GPU these ratios become 1.4, 1.9 and 3.2, respectively.           
\begin{table}
\centering
\caption{Accuracy and latency of the proposed detectors on Trash-ICRA19 dataset along with the number of epochs used for training. Architectural details are in Table \ref{tab:netsScaledFrom1-5}.}
 \begin{tabular}{ccccc|cc} 
 \hline
Architecture & \# epochs & \multicolumn{3}{c}{Accuracy (\%)} & \multicolumn{2}{c}{Latency (ms)} \\
& & AP & AP50 & AP75 & GPU & CPU \\   
\hline
D0(1-5) & 35 & 47.4 & 60.0 & 56.0 & 15 & 179\\ 
\hline 
D1(1-5) & 35 & 49.0 & 60.0 & 56.5 & 21 & 395\\
\hline 
D2(1-5) & 35 & 43.7 & 52.5 & 49.7 & 29 & 652\\ 
\hline
D3(1-5) & 35 & 49.6 & 56.9 & 55.5 & 48 & 1423\\
\hline
\end{tabular}
\label{tab:ResultsWithTrashDataset}
\end{table}

\begin{figure*}
\begin{subfigure}{0.24\textwidth}
  \centering
  \includegraphics[width=.99\linewidth]{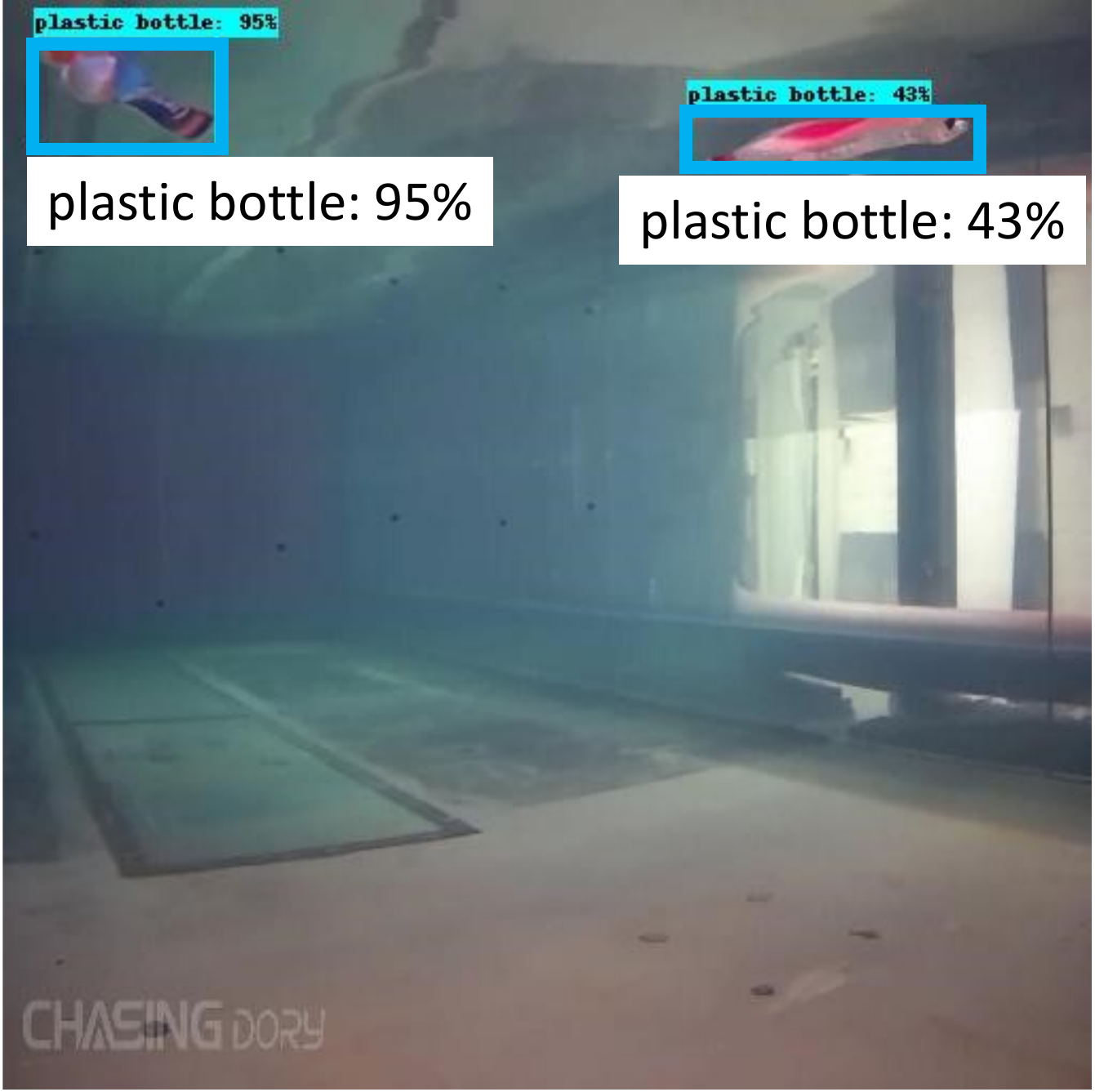}
  \caption{D0(1-5), $\tau = 40\%$}
  \label{fig:DebrisDetection3-D0}
  \vspace{5mm}
\end{subfigure}%
\begin{subfigure}{0.24\textwidth}
  \centering
  \includegraphics[width=.99\linewidth]{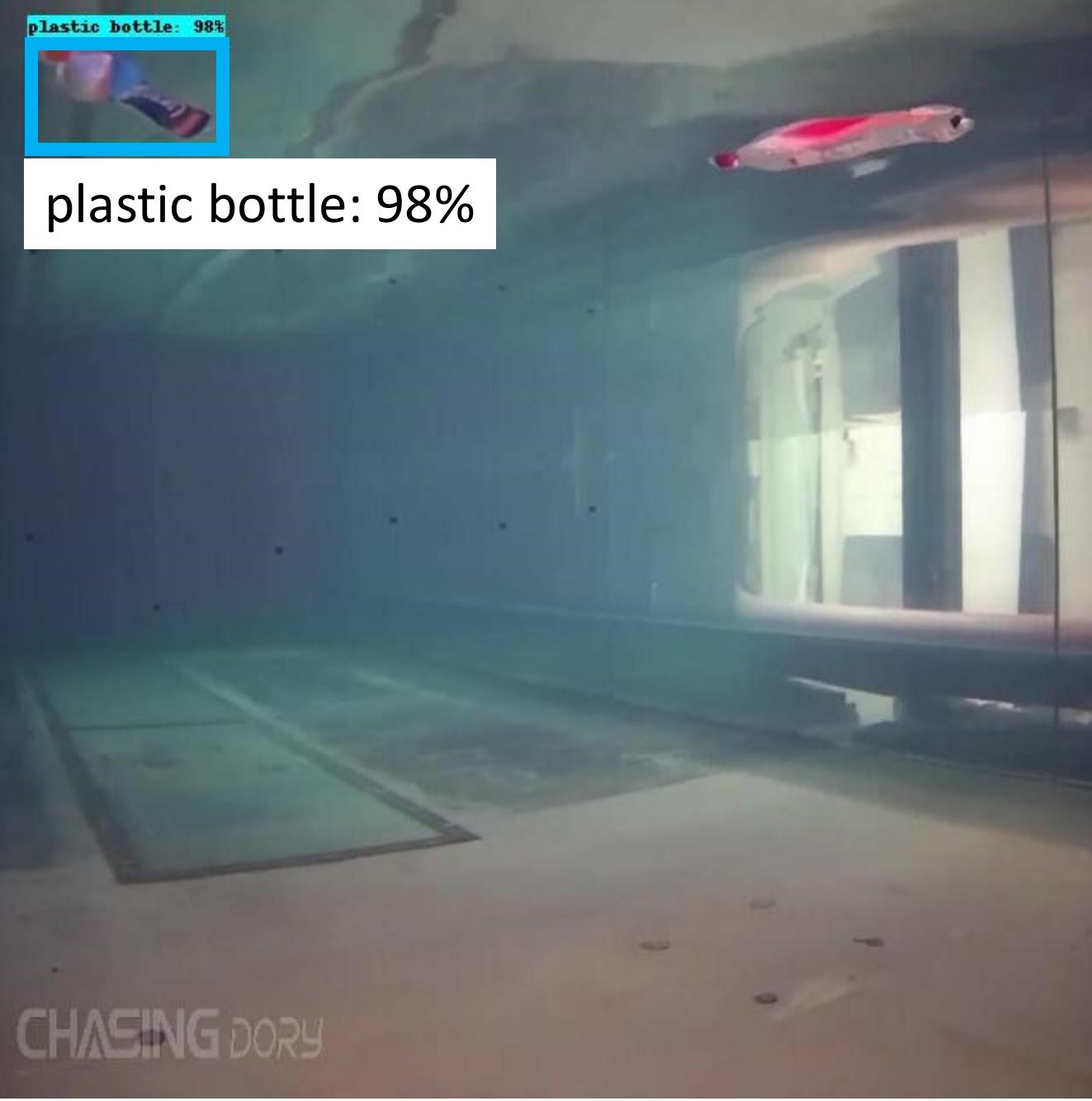}
  \caption{D1(1-5), $\tau = 95\%$}
  \label{fig:DebrisDetection3-D1}
  \vspace{5mm}
\end{subfigure}
\begin{subfigure}{0.24\textwidth}
  \centering
  \includegraphics[width=.99\linewidth]{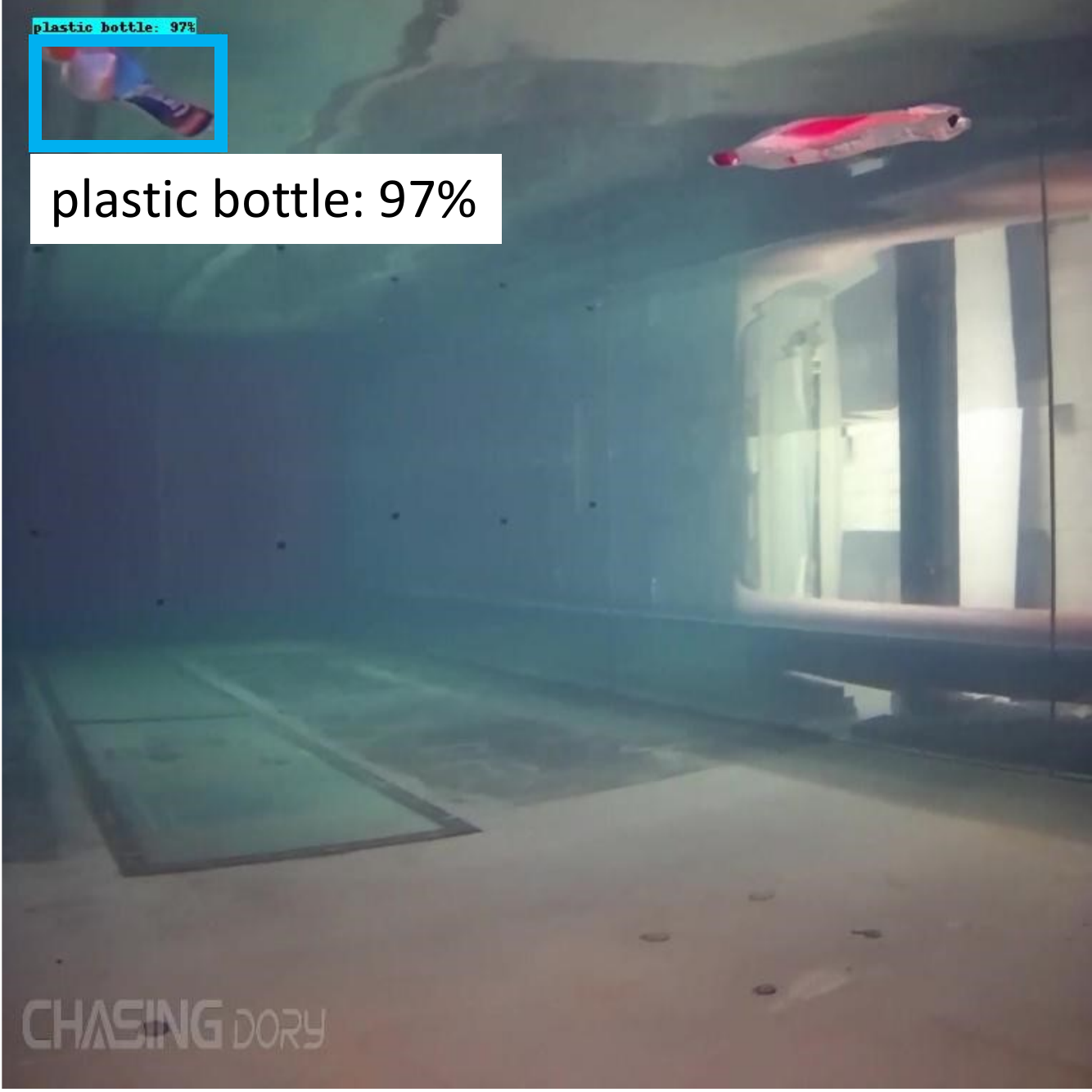}
  \caption{D2(1-5), $\tau = 95\%$}
  \label{fig:DebrisDetection3-D2}
  \vspace{5mm}
\end{subfigure}
\begin{subfigure}{0.24\textwidth}
  \centering
  \includegraphics[width=.99\linewidth]{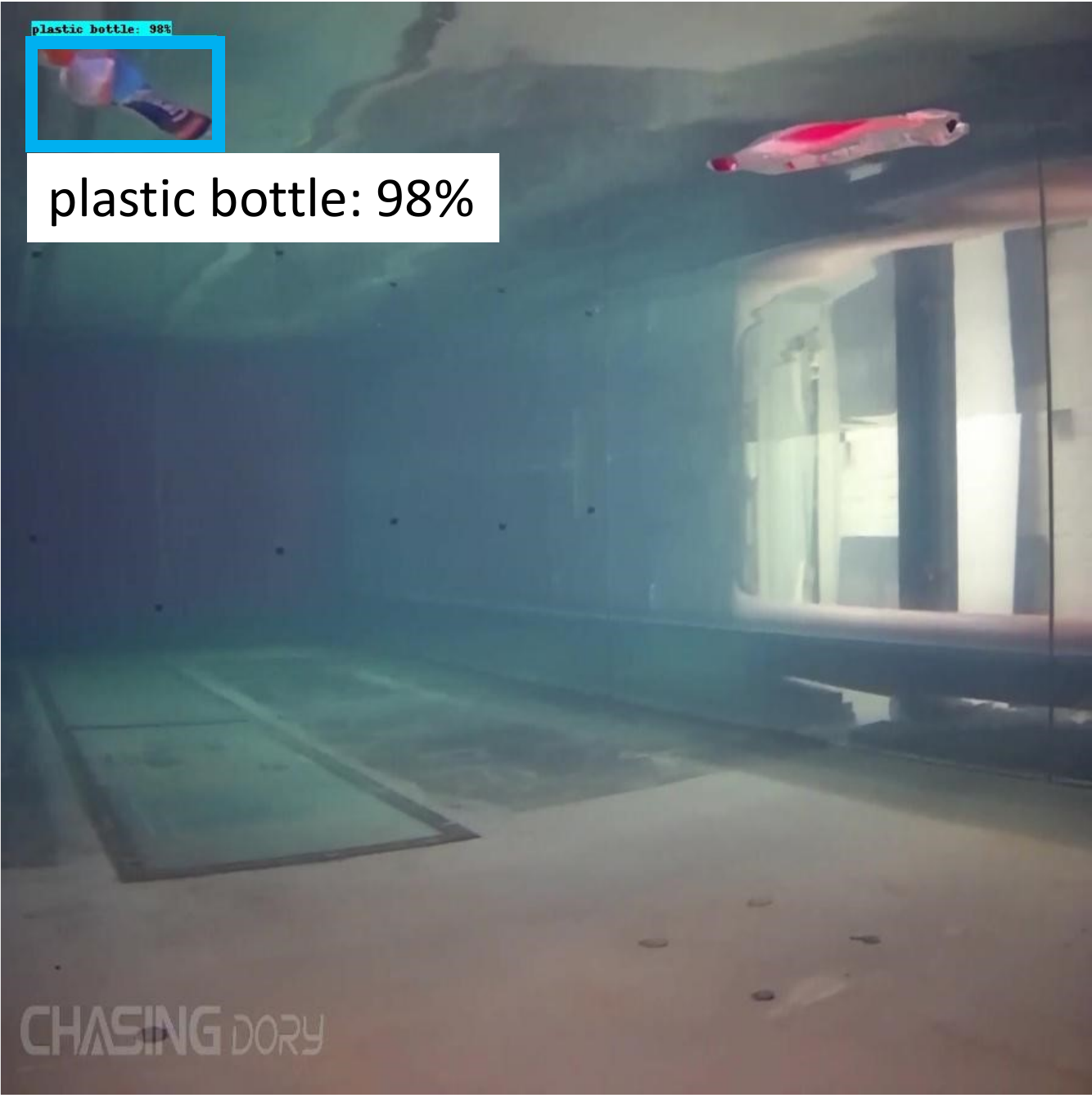}
  \caption{D3(1-5), $\tau = 95\%$}
  \label{fig:DebrisDetection3-D3}
  \vspace{5mm}
\end{subfigure} \\ 
\begin{subfigure}{0.24\textwidth}
  \centering
  \includegraphics[width=.99\linewidth]{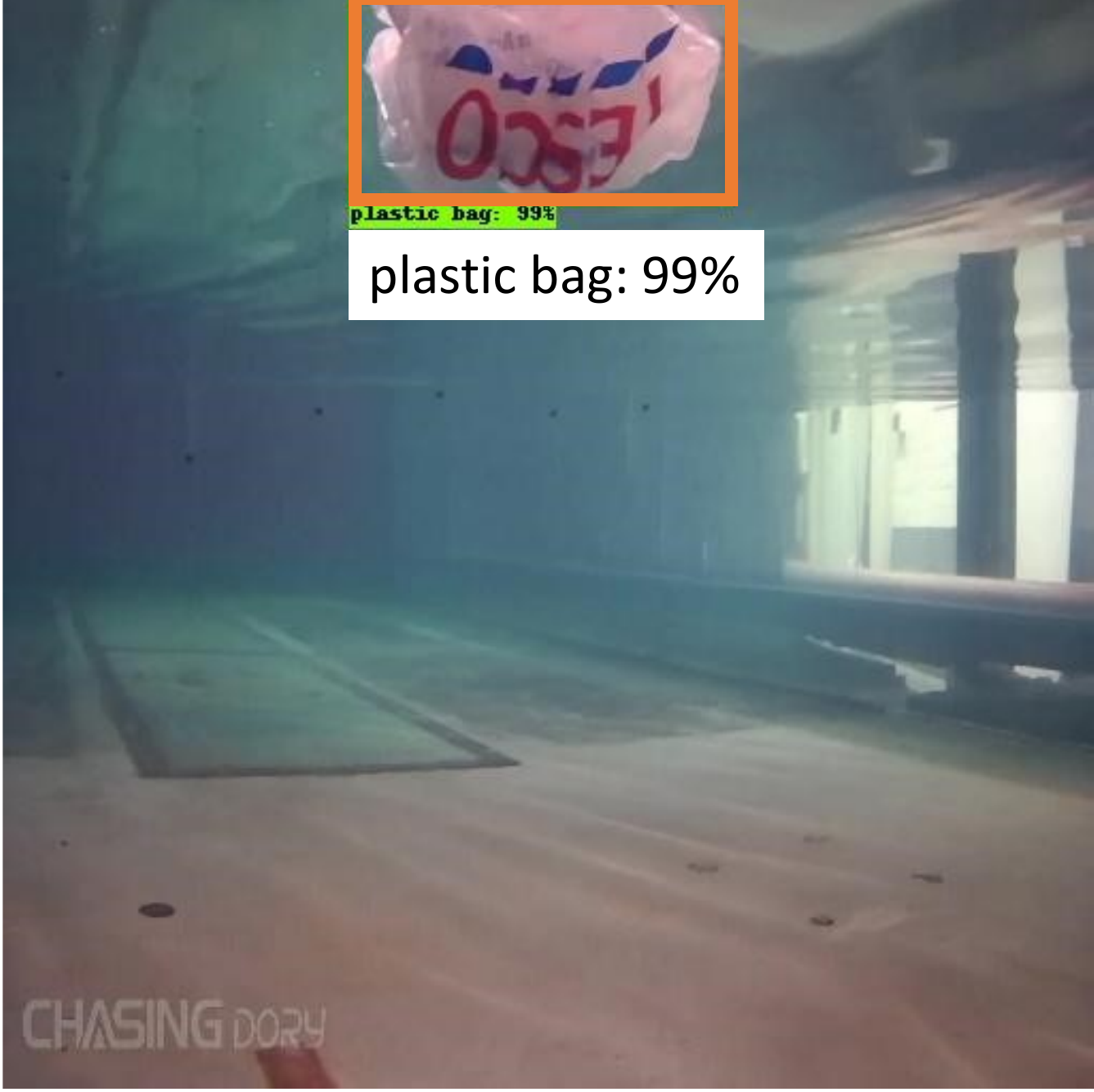}
  \caption{D0(1-5), $\tau = 95\%$}
  \label{fig:DebrisDetection4-D0}
\end{subfigure}%
\begin{subfigure}{0.24\textwidth}
  \centering
  \includegraphics[width=.99\linewidth]{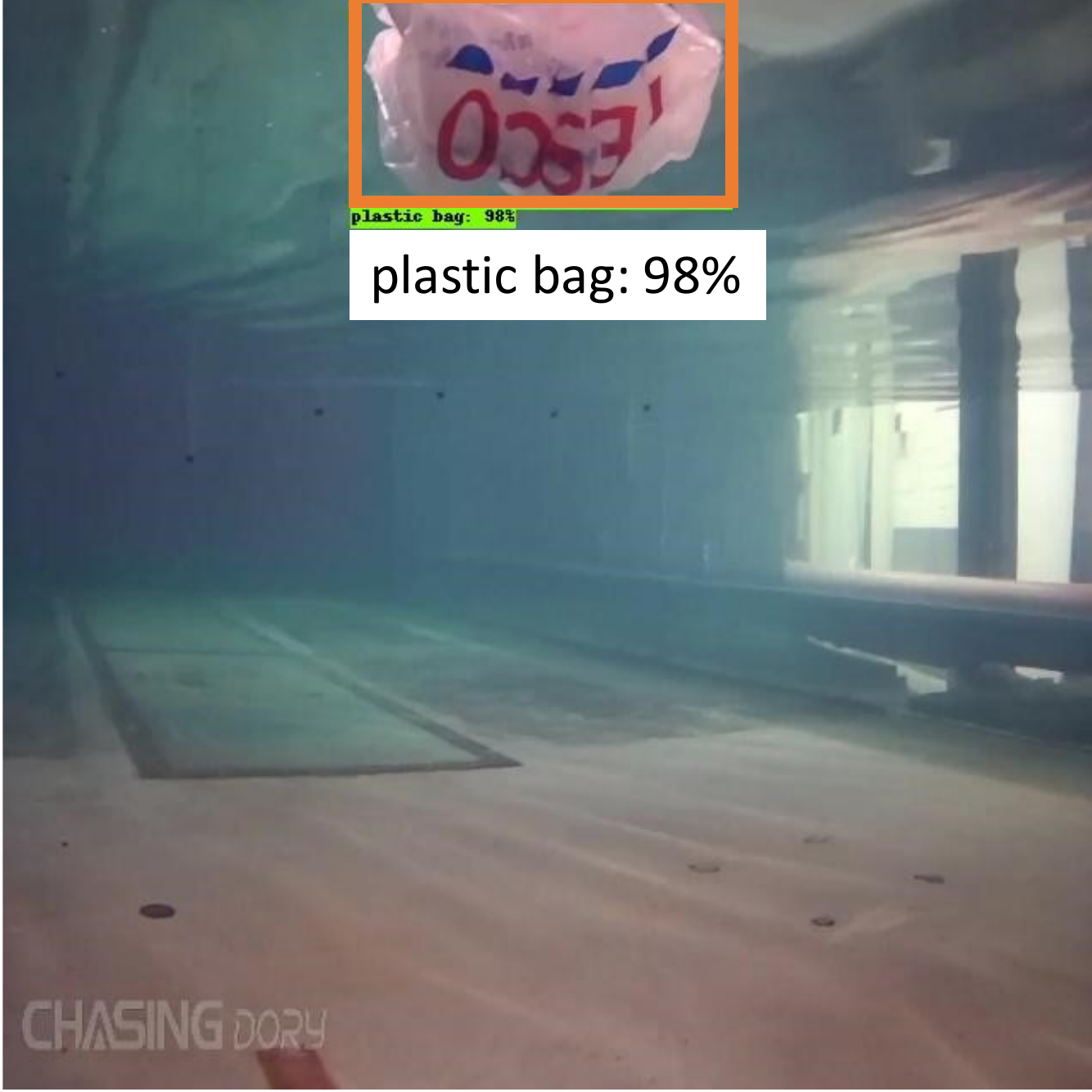}
  \caption{D1(1-5), $\tau = 95\%$}
  \label{fig:DebrisDetection4-D1}
\end{subfigure}
\begin{subfigure}{0.24\textwidth}
  \centering
  \includegraphics[width=.99\linewidth]{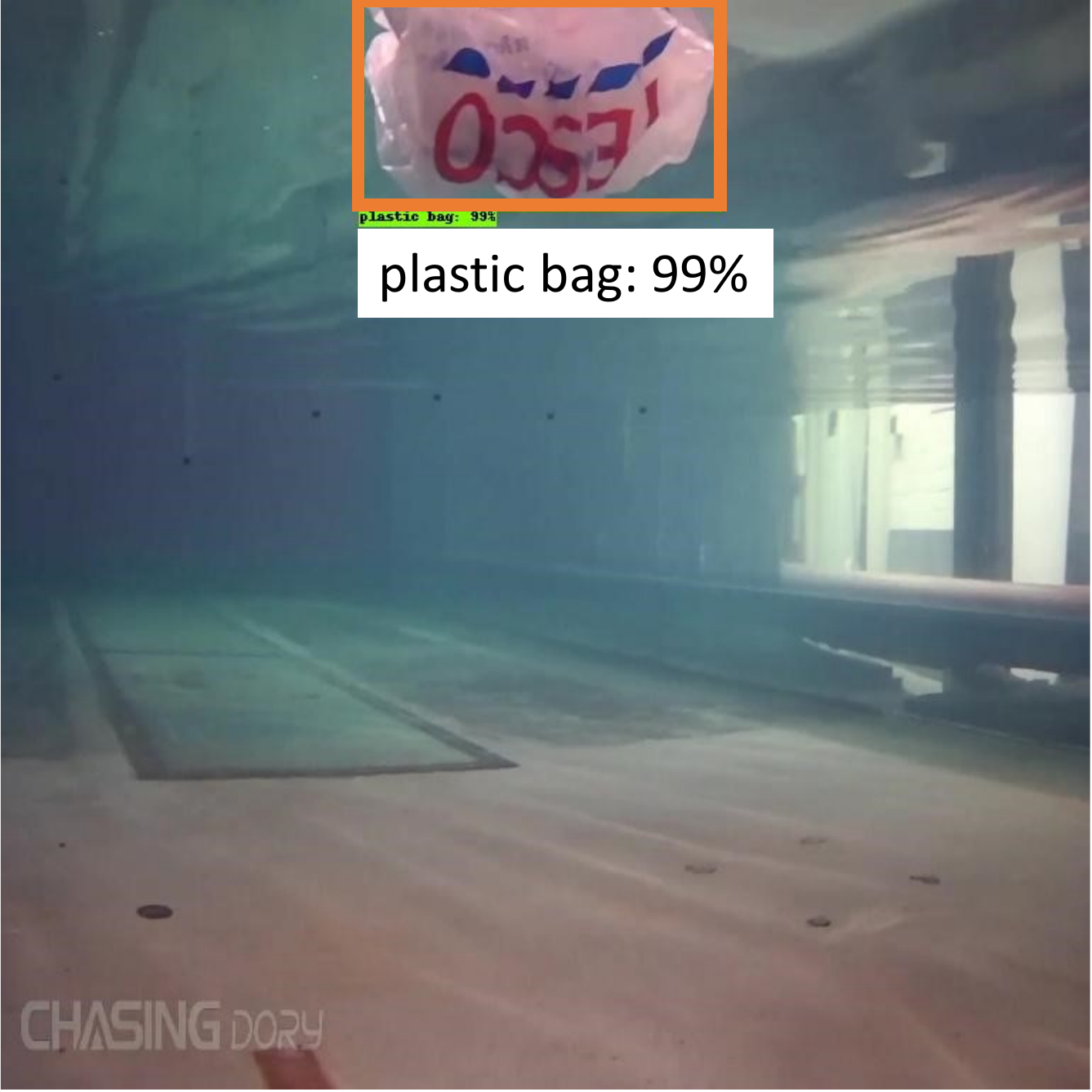}
  \caption{D2(1-5), $\tau = 95\%$}
  \label{fig:DebrisDetection4-D2}
\end{subfigure}
\begin{subfigure}{0.24\textwidth}
  \centering
  \includegraphics[width=.99\linewidth]{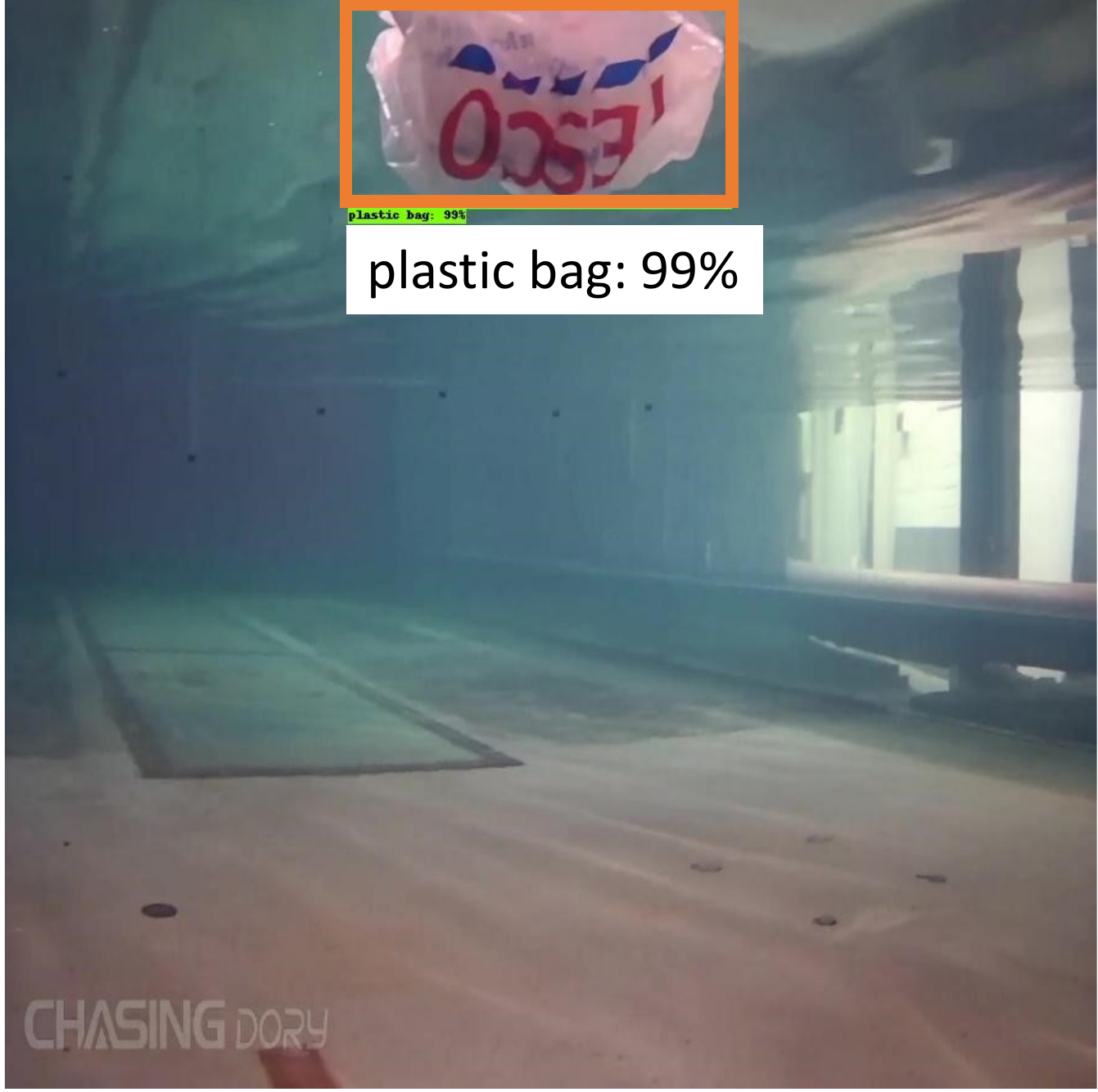}
  \caption{D3(1-5), $\tau = 95\%$}
  \label{fig:DebrisDetection4-D3}
\end{subfigure} 
\caption{Examples of detections on the WPBB dataset performed by D0(1-5) and its derivations; $\tau$ is the minimum confidence. The models producing these inferences are in Table \ref{tab:ResultsWithWPBBDataset}.}
\label{fig:DebrisDetectionsPBBW}
\end{figure*}

Since Trash-ICRA19 does not specify the type of plastic object detected, we created the in-water plastic bags and bottles (WPBB) dataset, which has 900 fully annotated images: 500 depict bags and 400 depict bottles. The images where generated by selecting the frames of videos recorded in a testing pool located in the David Keir Building at Queen's University Belfast. The videos of underwater waste items were recorded with a Chasing Dory drone as shown in the bottom left of Fig. \ref{fig:AutomationOfAUV}. We trained our EfficientDets on WPBB achieving approximately 15\% AP as detailed in Table \ref{tab:ResultsWithWPBBDataset}. Examples of detections are shown in Fig. \ref{fig:DebrisDetectionsPBBW} with different values of confidence threshold $\tau$, which is an hyperparameter meaning that detections with a confidence lower than $\tau$ are ignored.   
\begin{table}
\centering
\caption{Accuracy and latency of the proposed detectors on WPBB dataset along with the number of epochs used for training. Architectural details are in Table \ref{tab:netsScaledFrom1-5}.}
 \begin{tabular}{ccccc|cc} 
 \hline
Architecture & \# epochs & \multicolumn{3}{c}{Accuracy (\%)} & \multicolumn{2}{c}{Latency (ms)} \\
& & AP & AP50 & AP75 & GPU & CPU \\   
\hline
D0(1-5) & 50 & 15.1 & 17.3 & 16.8 & 15 & 179\\ 
\hline 
D1(1-5) & 40 & 14.8 & 17.3 & 16.8 & 21 & 395\\
\hline 
D2(1-5) & 40 & 14.9 & 17.3 & 17.3 & 29 & 652\\ 
\hline
D3(1-5) & 36 & 15.2 & 17.3 & 16.8 & 48 & 1423\\
\hline
\end{tabular}
\label{tab:ResultsWithWPBBDataset}
\end{table}

\subsection{Low-Light Underwater Conditions}\label{sub:UIE}
\begin{figure}
\includegraphics[width=0.43\textwidth]{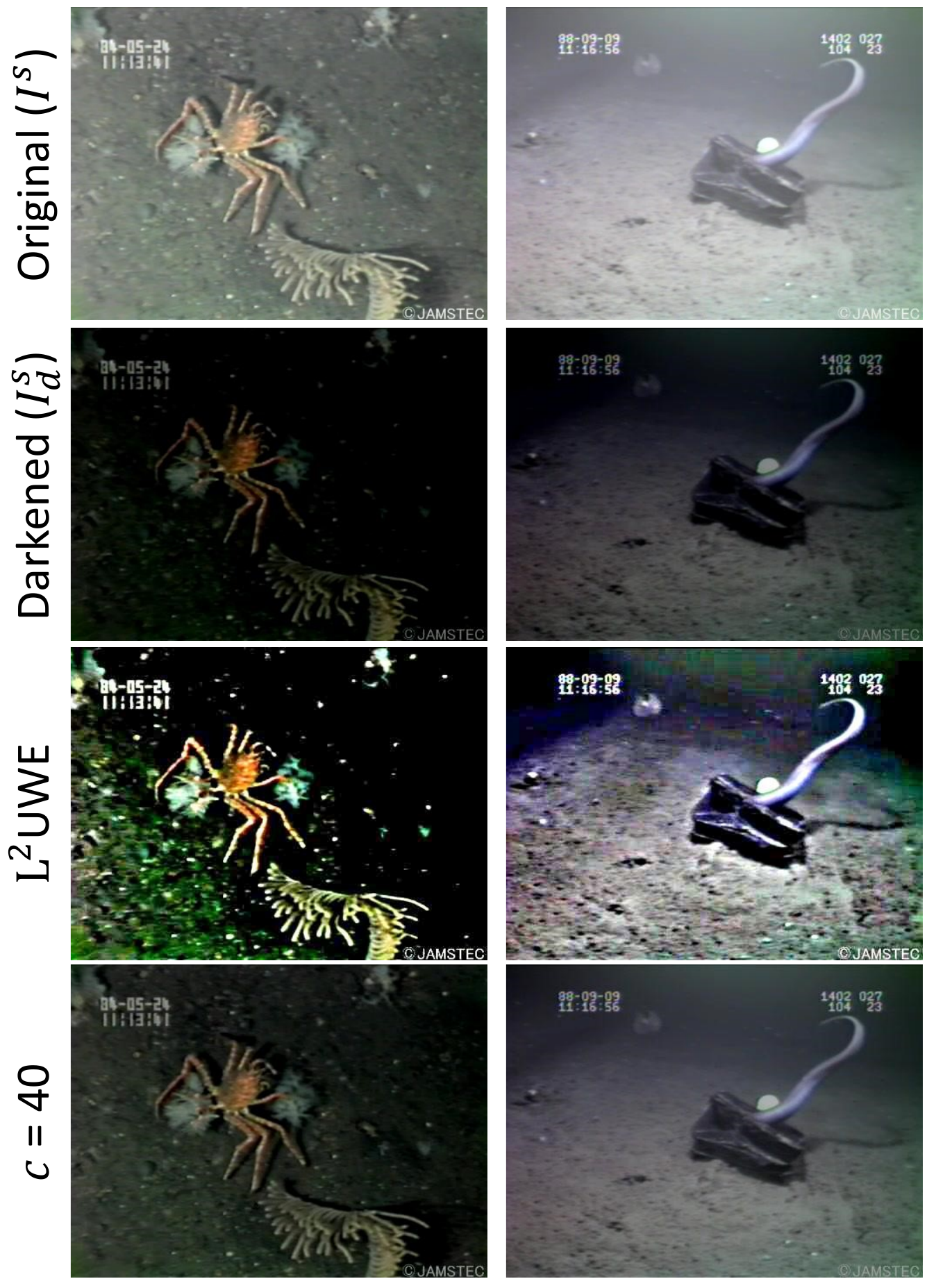}
\centering
\caption{Examples of images used to evaluate the performance of low-light underwater image enhancement strategies as pre-processing step before an object detector as in Fig. \ref{fig:AutomationOfAUV}.}
\label{fig:EnhancedImages}
\end{figure}

Water depth or the use of AUVs at night time could lead to low-light underwater images recorded by the on-board RGB camera. In this subsection, we investigate how low-light conditions could affect object detection and how the low-light image enhancement method proposed in \cite{L2UWE} works when used as pre-processing step before an object detector as described in Fig. \ref{fig:AutomationOfAUV}. To simulate low-light conditions, we modified the first 300 samples of Trash-ICRA19 by simply subtracting a constant value from all the pixels of each image/sample, that is,
\begin{equation}
I^{s}_{d} = I^{s} - 120,
\end{equation}
where $I^{s}_{d}$ is the darkened version of the $s$-th image $I^{s}$ of Trash-ICRA19 and $s = {1, 2, \dots, 299, 300}$. Two examples of the outcome are the first two rows of Fig. \ref{fig:EnhancedImages}. Then, we considered four scenarios to compare the inference accuracy of the detectors trained previously on Trash-ICRA19 (see Table \ref{tab:ResultsWithTrashDataset}):
\begin{enumerate}
\item{No any low-light image enhancement method is used, i.e. the darkened images are the input of the detectors.}
\item{The low-light image enhancement method proposed in \cite{L2UWE} (named ``$\text{L}^2$UWE'') improves the darkened images, which are then given as input to a detector.}
\item{The darkened images are enhanced by adding a constant value $c = 40$ to all the pixels, i.e. $I^{s}_{e} = I^{s}_{d} + c$, where $I^{s}_{e}$ is the enhanced version of $I^{s}_{d}$; then, $I^{s}_{e}$ is given as input to a detector.}
\item{The same as the previous scenario, but using $c = 80$.}
\end{enumerate}
A qualitative assessment of the outcome of $\text{L}^2$UWE and $c = 40$ is shown in Fig. \ref{fig:EnhancedImages} where it is visible the difference between the non-uniform and the uniform increase of light given by the former and the latter, respectively.      

The result considering the four scenarios are in Table \ref{tab:withEnhancements} along with the computational time required by each enhancement method to improve a single image.   
\begin{table}
\centering
\caption{Detection accuracy without and with enhancement methods along with the latency of each enhancement method, i.e. the computational time to enhance a single image.}
 \begin{tabular}{ccccc|c} 
 \hline
Detector & Enh. method & AP & AP50 & AP75 & CPU latency (s)\\  
\hline
\multirow{4}{*}{D0(1-5)} & None & 31.4 & 44.9 & 36.8 & \multirow{10}{*}{$\text{L}^2$UWE}\\
& $\text{L}^2$UWE & 39.4 & 49.2 & 44.4 & \multirow{10}{*}{11.1}\\
& $c$ = 40 & 50.5 & 62.8 & 58.7 & \multirow{18}{*}{$c$ = constant}\\ 
& $c$ = 80 & \textbf{56.1} & \textbf{70.5} & \textbf{66.4} & \multirow{18}{*}{\textbf{0.02}}\\  
\cline{1-5}
\multirow{4}{*}{D1(1-5)} & None & 5.8 & 19.0 & 3.7 &\\
& $\text{L}^2$UWE & 38.9 & 54.6 & 43.9 &\\
& $c$ = 40 & 41.3 & 56.0 & 47.9 &\\ 
& $c$ = 80 & \textbf{51.3} & \textbf{73.9} & \textbf{61.1} &\\
\cline{1-5}   
\multirow{4}{*}{D2(1-5)} & None & 7.2 & 12.6 & 6.6 &\\
& $\text{L}^2$UWE & 51.3 & 65.3 & 57.5 &\\
& $c$ = 40 & 54.8 & 70.5 & 61.3 &\\ 
& $c$ = 80 & \textbf{59.0} & \textbf{71.9} & \textbf{68.2} &\\
\cline{1-5}   
\multirow{4}{*}{D3(1-5)} & None & 20.5 & 28.9 & 22.3 &\\
& $\text{L}^2$UWE & 57.8 & \textbf{70.9} & 66.6 &\\
& $c$ = 40 & 47.7 & 62.2 & 50.4 &\\ 
& $c$ = 80 & \textbf{61.4} & 69.4 & \textbf{68.3} &\\     
 \hline
 \end{tabular}
 \label{tab:withEnhancements}
\end{table}
As mentioned in the Introduction section, the first of the two questions we aim to answer is: ``Is it more efficient scaling-up the object detector size (e.g. from D0(1-5) to D3(1-5)) or adding an underwater image enhancement method before the smallest detector D0(1-5)?''. Table \ref{tab:withEnhancements} suggests that D0(1-5) is the most accurate detector if no enhancement methods are used (31.4\% AP with D0, 5.8\% AP with D1, 7.8\% AP with D2 and 20.5\% AP with D3), which is also the smallest. To improve its accuracy, an enhancement method could be used: $\text{L}^2$UWE provides an improvement of 8\% AP at the cost of 11.1 seconds to enhance a single image, $c = 40$ provides an improvement of 19.1\% AP at the cost of 20 milliseconds per image and $c = 80$ an improvement of 24.7\% AP at the cost of 20 milliseconds per image; hence, the total inference times on a CPU are 0.179 s + 11.1 s = 11.279 seconds with $\text{L}^2$UWE and 0.179 s + 0.020 s = 0.199 seconds with $c$-based methods. To further increase the accuracy, D0(1-5) could be scaled-up to D3(1-5): this results in a detector 7.9 times slower on a CPU (179 milliseconds with D0 vs. 1423 milliseconds with D3) and 3.2 times slower on a GPU, hence the total inference time per image with D3 becomes 1.423 s + 11.1 s = 12.523 seconds with $\text{L}^2$UWE and 1.423 s + 0.020 s = 1.443 seconds with $c$-based methods on a CPU, whereas the evaluation on a GPU requires the implementation of the enhancement methods on a GPU, which is currently not available. However, D3 can reach 61.4\% AP accuracy, which is the highest, but only 5.3\% AP higher than D0 (with 56.1\% AP). Note that the number of epochs chosen in this study to train the models can be increased and this should result in a larger difference in accuracy between D0 and D3. In conclusion, for real-time underwater marine detection using a CPU, the suggested detectors are D0(1-5) and D1(1-5) if trained for more epochs in order to leverage its bigger size compared to D0. If a GPU is available on-board, D2(1-5) and D3(1-5) could also be considered for real-time detection provided that they are trained with more epochs in order to fully exploit their potential. In case a low-light image enhancement is needed, $\text{L}^2$UWE is too computationally demanding for real-time applications (this is the answer to the second question we asked in the Introduction section). A more efficient approach is to tune the value of $c$ off-line using genetic algorithms or particle swarm optimization and subsequently use it for low-light real-time marine debris detection within an automation pipeline as the one in Fig. \ref{fig:AutomationOfAUV}.

\section{CONCLUSION}
\label{sec:Conclusions}
As marine debris is an increasing concern for both human and wildlife health, AUVs could be used for debris removal. In this letter, the efficiency of debris detection is addressed by improving the efficiency of EfficientDets, creating and publishing a fully annotated dataset for in-water plastic bags and bottles, and finally considering low-light underwater conditions. Future work will consist of implementing the detectors on an AUV and integrate the detection outputs with the control system of both vehicle and gripper for debris removal.

\addtolength{\textheight}{-4cm}   





\end{document}